\documentclass[acmsmall]{acmart}

\AtBeginDocument{%
	\providecommand\BibTeX{{%
			\normalfont B\kern-0.5em{\scshape i\kern-0.25em b}\kern-0.8em\TeX}}}

\copyrightyear{2021}

\usepackage{lineno,hyperref}
\usepackage{amsmath}
\usepackage{amsfonts}
\usepackage{algorithm}
\usepackage[noend]{algpseudocode}
\usepackage{enumitem}
\usepackage{graphicx}

\begin{document}
	
	\title{Towards Implementing Energy-aware Data-driven Intelligence for Smart Health Applications on Mobile Platforms}
	
	\author{G. Dumindu Samaraweera}
	\email{samaraweera@usf.edu}
	\orcid{0000-0003-4097-5585}
	\author{Hung Nguyen}
	\email{nsh@usf.edu}
	\author{Hadi Zanddizari}
	\email{hadiz@usf.edu}
	\author{Behnam Zeinali}
	\email{behnamz@usf.edu}
	\author{J. Morris Chang}
	\email{chang5@usf.edu}
	\affiliation{%
		\institution{University of South Florida}
		\streetaddress{Department of Electrical Engineering, 4202 E. Fowler Avenue}
		\city{Tampa}
		\state{Florida}
		\postcode{33620}
	}

	\renewcommand{\shortauthors}{Samaraweera, et al.}
	
	\begin{abstract}
		Recent breakthrough technological progressions of powerful mobile computing resources such as low-cost mobile GPUs along with cutting-edge, open-source software architectures have enabled high-performance deep learning on mobile platforms. These advancements have revolutionized the capabilities of today's mobile applications in different dimensions to perform data-driven intelligence locally, particularly for smart health applications. Unlike traditional machine learning (ML) architectures, modern on-device deep learning frameworks are proficient in utilizing computing resources in mobile platforms seamlessly, in terms of producing highly accurate results in less inference time. However, on the flip side, energy resources in a mobile device are typically limited. Hence, whenever a complex Deep Neural Network (DNN) architecture is fed into the on-device deep learning framework, while it achieves high prediction accuracy (and performance), it also urges huge energy demands during the runtime. Therefore, managing these resources efficiently within the spectrum of performance and energy efficiency is the newest challenge for any mobile application featuring data-driven intelligence beyond experimental evaluations. In this paper, first, we provide a timely review of recent advancements in on-device deep learning while empirically evaluating the performance metrics of current state-of-the-art ML architectures and conventional ML approaches with the emphasis given on energy characteristics by deploying them on a smart health application. With that, we are introducing a new framework through an energy-aware, adaptive model comprehension and realization (EAMCR) approach that can be utilized to make more robust and efficient inference decisions based on the available computing/energy resources in the mobile device during the runtime.
	\end{abstract}

	\ccsdesc[500]{Computing methodologies~Machine learning}
	\ccsdesc[500]{Computing methodologies~Neural networks}
	\ccsdesc[500]{Computing methodologies~Object detection}
	\ccsdesc[500]{Computing methodologies~Image segmentation}

	\keywords{Mobile Deep Learning, Machine Learning, Smart Health, Energy-aware, Mobile Computing Performance.}

	\maketitle
	
	\section{Introduction}
	Due to the recent most exciting advancements of deep learning, the performance of conventional machine learning algorithms has beaten in numerous ways. These advancements have revolutionized the capabilities of most data-driven applications that we use for our day-to-day work. Eventually, once these applications become vital, people have explored the possibilities to bring machine learning into mobile devices. One of the key contributors to bring powerful deep neural networks (DNNs) into mobile platforms comes from the advancements in mobile computing resources. In fact, vast variety of mobile devices available in the market today are equipped with mobile GPUs and powerful CPUs which can be efficiently utilized in deep learning tasks \cite{ignatov2019ai}. On the other hand, recent most breakthrough developments in open source on-device deep learning frameworks (e.g. TensorFlow Lite \cite{tensorflowlite}, PyTorch Mobile \cite{pytorchmobile}) and APIs (e.g. Android NNAPI \cite{nnapi2019}) have paved the pathway to bring modern hardware into existing mobile platforms. Thereby, today, on-device machine learning can be successfully deployed in many application scenarios such as data classification, object recognition and detection, natural language processing (NLP) across a vast variety of domains such as healthcare, marketing, finance and so on. 
	
	Nevertheless, deploying a DNN architecture on a mobile platform carries a couple of unprecedented challenges. These sophisticated DNN architectures perform well on modern hardware and software platforms; however, on the flip side, it urges huge resource demands during the runtime which eventually to be fulfilled by a constrained set of energy resources accessible to the mobile device. Hence, whilst the convenience of modern computing resources has made on-device deep learning an increasingly popular and promising solution for today’s data driven applications, there is a clear-cut tradeoff that needs to be addressed within the spectrum of performance/accuracy and energy utilization. On the other hand, for these applications to be realizable in industry level deployments, energy efficiency is a vital factor that governs the survival of the application for the long run. While different studies have been discussed in the literature covering various aspects in terms of deploying on-device deep learning on mobile platforms \cite{deng2019deep}, \cite{wang2018deep}, \cite{guo2018cloud}, \cite{ran2017delivering}, \cite{ogden2018modi}, we intend to devise this problem in a different perspective. Our objective in this study is to first analyze modern DNN architectures in terms of accuracy, performance and energy characteristics toward deploying them on practical applications (such as smart health/telemedicine). Secondly, the possibility of blending modern powerful hardware/software architectures into mobile platforms is explored by emphasizing the challenges of handling mobile deep learning frameworks energy efficiently. And thirdly, we intend to put forward a solution to these challenges by introducing an energy-aware DNN based intelligent framework toward the practical end.
	
	Over the past, both the academia and industry have explored the possibility of how to deploy DNNs efficiently on edge devices, particularly on mobile platforms \cite{deng2019deep}, \cite{wang2018deep}, \cite{guo2018cloud}, \cite{ran2017delivering}, \cite{ogden2018modi}. However, it is noteworthy that most of these solutions have overlooked the importance of incorporating energy awareness of the DNN architectures into their design, which we believe to be the crucial factor for modern data-driven applications in terms of them to be realizable in practice. While there are many application scenarios for on-device deep learning, healthcare applications, particularly smart health, has a potentially profound impact over others \cite{deng2019deep}. With various built-in and external sensors, mobile devices can be utilized as medical devices in the field to extract physical and psychological data from individuals to perform numerous diagnosis types. In addition, deep neural networks are useful in making fast, powerful and accurate predictions for diseases. Hence they can be utilized to take timely and predictive measures. On the other hand, lack of labeled data is a major concern in healthcare applications; however, deep learning models can be optimized using different data pre-processing (e.g. augmentation, object detection, segmentation) and transfer learning techniques to perform better in such circumstances \cite{ran2017delivering}, \cite{misbhauddin2019initial}. Moreover, deep learning models can be operated even with low quality data, which makes these applications more realistic for smart health in practice. Therefore, in this article, we choose smart health/telemedicine in a resource contested environment as an example use case scenario to showcase the energy requirements/characteristics of deep learning models, and to propose an energy-aware solution for on-device deep-learning-powered mobile applications.
	
	In essence, deep learning typically involves a sophisticated network architecture along with a great number of parameters that govern the deep neural network itself. Hence, in reality the resultant model can be very different in size which can introduce huge resource demands during the time of model loading and execution. In addition, the complexity of the inference task usually depends on the size of the input data (e.g. resolution of the image in an image classification task) and the deep learning model in use. Whenever the complexity is high, it consumes more energy, attributing to high memory and CPU usage, and resulting higher latency during the time of inference. Apart from the aforementioned motives, on the other hand, different mobile devices have different resource capabilities. Nowadays, almost every smartphone has some form of an integrated mobile GPU (such as Qualcomm Adreno \cite{adrenoGPU}, Imagination PowerVR \cite{powerVRGPU}, etc.) to enhance the user experience through high performance graphics which can also be comprehended to deep learning tasks. However, sometimes due to the hardware specialization, it is quite challenging for some of the accelerators (e.g. GPU and Android NNAPI\cite{nnapi2019}) to perform well on even most recent deep learning models such as \cite{effnetlite2020} due to various reasons including architectural limitations. In addition, in most scenarios, quantization is widely used to reduce the model size and then to improve the accelerator latency. But, it usually requires complex and quantization-aware training procedure which is quite challenging to realize in practice. 
	
	In this article, we provide a timely review of the current challenges and the way-forward towards deploying cutting-edge deep learning on mobile platforms by choosing smart health/telemedicine as an example use case in a resource contested environment. Our solution also consists of a novel energy-aware approach focusing on how to utilize limited software/hardware resources in a mobile device to achieve on-device deep learning with high accuracy and efficiency. At first, we conduct an extensive set of experiments to measure the capability/suitability of latest cutting-edge deep neural networks having more emphasis on energy efficiency for on-device deep learning beyond the conventional machine learning algorithms. Then, we analyze the performance of each network architecture for three different healthcare application scenarios: 1) eardrum abnormality diagnosis, 2) fingernail abnormality diagnosis and 3) skin lesion diagnosis by categorizing them in terms of model accuracy, latency and energy efficiency. Secondly, we introduce our energy-aware, adaptive model comprehension and realization (EAMCR) approach where we put forward all these different constraints when deploying DNNs on mobile platforms to design a robust and efficient smart health application to the practical end with a higher level of accuracy while making our solution deployable even in uncontrolled clinical conditions. For the best of our knowledge, we are the first to investigate energy awareness of the current/latest state-of-the-art DNNs for on-device deep-learning-powered smart health/telemedicine applications with an adaptive model comprehension and realization approach during the runtime.
	
	The remaining sections of the paper are structured as follows: Section 2 reviews the background of mobile based machine learning approaches and the related work particularly in healthcare applications. Section 3 presents the details of image analysis and data acquisition methods for image pre-processing while Section 4 discusses the implementation of mobile deep learning framework. An analysis along with a performance evaluation is given in the Section 5 including our proposed energy-aware and adaptive model realization framework for resource contested environments. Finally, the paper concludes with Section 6 by highlighting the challenges and key considerations for the future work.
	
	\vspace*{-3mm}
\section{Background and Related Work}
Designing an efficient framework for on-device deep learning and use of those architectures for healthcare applications is in the discussion for quite sometime. However, deploying DNNs on mobile devices has gained the popularity during the recent past along with the evolution of more efficient neural network architectures and the powerful computing resources available in modern smartphones. In this section, we provide background on different deep learning architectures tailored for mobile platforms and the related work in terms of utilizing those architectures in smart health/telemedicine applications.


\vspace*{-3mm}
\subsection{Deep Learning on Mobile Platforms}
It is clear that training a deep neural network is a much resource intensive task; thereby, DNNs are often trained using powerful CPU/GPU-based servers over a long period of time. But, once the model is trained, the inference can be done with comparatively less amount of computing resources including edge devices (e.g. mobile device, IoT). Typically, these trained models host in a cloud-based server and extend the accessibility for remote applications through web-based APIs. Despite the fact that simplicity of this approach, it has several drawbacks. At first, it requires the mobile application most of the time to be connected with the server which is not always realistic in practice. On the other hand, it raises deeper concerns on data privacy of collected information \cite{al2019privacy}. Due to these practical challenges in traditional ML deployment approaches, and the availability of powerful computing resources in the present-day mobile devices, on-device deep learning has gained the increasing popularity among today's data-driven applications.

Over the past, a number of popular and competitive deep learning frameworks such as TensorFlow Lite by Google \cite{tensorflowlite}, PyTorch Mobile (formerly known as Caffe2) by Facebook \cite{pytorchmobile}, and Core ML by Apple \cite{coreml} have evolved with the capabilities that can leverage deep learning on mobile devices. The mobile applications powered by these popular platforms can be broadly classified into two classes: 1) cloud-assisted deep learning solutions and 2) on-device deep learning solutions. While the quickest way to adopt deep learning capability on a mobile device is to take advantage of the existing cloud assisted AI platforms/APIs such as Google Cloud Vision, our primary focus of this paper is to utilize the capabilities of on-device deep learning specially in healthcare applications.

\subsubsection*{\textbf{ML Model Optimization Approaches}}
Apart from the deep learning frameworks, moving the original inference task to the mobile device is not really straight forward due to many reasons. There are various hardware constraints (such as the limitations in CPU, memory and battery capacity) making them ill-suited for on-device inference in many cases. Hence, people have looked into the ML model optimization techniques such that to redesign the models to use with less powerful hardware. Broadly, these model optimization techniques can be classified into two namely; 1) model size reduction and 2) latency reduction. Smaller models have couple of different benefits including less storage space requirements, smaller download size and less memory usage which helps better performance and stability \cite{tfliteoptimize2020}. On the other hand, optimization can also be applied in terms of reducing the amount of computation to run the inference, resulting lower latency which has a greater impact on power consumption. Quantization is one of the popular approaches in this category to compress the existing models. In essence, it is a process of approximating floating-point numbers in a neural network by a low bit width numbers so that it can reduce the memory requirement and the computational cost, hence improve the latency \cite{lin2016fixed}. 

\subsubsection*{\textbf{Deep Learning Architectures Tailored for Mobile/Edge Devices}}
There has been a lot of research on deep learning architectures tailored for mobile/edge devices. In 2016, Iandola et al. proposed SqueezeNet \cite{iandola2016squeezenet} a small DNN architecture that can achieve AlexNet-level \cite{krizhevsky2012imagenet} accuracy on ImageNet dataset with 50x fewer parameters. Later, Howard et al. \cite{howard2017mobilenets} proposed another class of efficient CNN models called MobileNets \cite{howard2017mobilenets} for mobile and embedded vision applications. Recently, in 2019, Tan et al. proposed a family of CNN models called EfficientNets \cite{tan2019efficientnet}, with better accuracy and efficiency than previous convolutional neural networks which can also be utilized in mobile devices. It is noteworthy that most of these tailored DNNs are focused on reducing the model parameters, in order for them to fit into mobile platforms. However, it is also important for these architectures to look into the energy awareness, which we believe to be one of the critical factors to consider in practical deployments. 

\subsubsection*{\textbf{Effect of Hardware Acceleration}}
With the evolution of complex network structures, deep learning models require huge computational resources and hence running them on CPUs was nearly infeasible from both the performance and power efficiency perspective. In 2015 Qualcomm made the first attempt to accelerate deep learning models on mobile GPUs and DSPs \cite{ignatov2019ai}. Over the past few years, various other manufacturers also developed different AI processing units in a rapid phase making them closer to the results of powerful desktop hardware.  

Despite these advancements of different deep learning architectures and hardware platforms, yet, most of the existing studies \cite{latifi2016cnndroid}, \cite{lane2016deepx}, \cite{kim2019mulayer} have not considered how to blend these hardware accelerators with cutting-edge on-device DNN architectures toward deploying energy-aware solutions. This absence of comprehensive study on energy awareness of deep learning algorithms has motivated us on finding a solution toward designing a framework that can be deployed on mobile platforms toward the practical-end of healthcare applications. 

\vspace*{-3mm}
\subsection{Use of Deep Learning in Healthcare Applications}
It has been noticed that a majority of medical diagnosis systems are based on image data; however, in many situations, detection of diseases from these images is a difficult task even for medical professionals because of the variability in appearance. Hence, today's advanced Convolutional Neural Networks (CNNs) can be adaptively used in healthcare applications to obtain more accurate and reliable disease predictions. A set of sample images with a certain disease can be trained for a image recognition/classification task and once the model is trained, a simple application can be deployed to diagnose new and unseen images to examine whether those images are with a certain disease. Typically, the accuracy of these CNN models is undoubtedly high and therefore, during the recent past, greater portion of applications in healthcare domain has made effective use of machine learning in some way or the other \cite{misbhauddin2019initial}. 

Broadly, the mobile healthcare systems powered by deep neural networks can be classified into three classes. The first category is cloud-assisted mobile healthcare applications where the deep learning model is deployed in the cloud-based server and mobile application is serving as the front-end interface. However, the efficiency of these solutions are heavily relying on the performance of underlying communication infrastructure hence not well-suited for delay-sensitive medical services \cite{muhammed2018ubehealth} and resource contested environments. Moreover, most of these solutions are utilizing multi levels of Internet of Things (IoT) devices/sensors to collect information and then those information are processed and analyzed in the cloud for powerful predictions \cite{habibzadeh2019survey}. These applications have been used in different disease diagnosis scenarios such as early detection of heart diseases \cite{kumar2018novel} and sustainable health monitoring tools \cite{chen2016smart}, \cite{hossain2016cloud}. The second category is looking at the healthcare solutions powered by edge-computing or fog-computing to reduce the problem of high network utilization in the first category \cite{gia2019exploiting}. To that end, multiple studies have also been carried out \cite{muhammed2018ubehealth}, \cite{pace2018edge} to explore the possibility of fog-computing in healthcare applications. 

The final and third category is focused on on-device deep learning based healthcare applications. Typically, these on-device applications are enabled with optimized light-weight machine learning models and comparatively lower accuracy compared to other design choices. However, its main advantage is it can work very well under resource contested environments even without needing to use a network connection. Recently several studies have been proposed to identify skin cancer through mobile based deep learning approaches \cite{kalwa2019skin}, \cite{ech2019deep}, \cite{do2018accessible}. It is noteworthy that, some of these works are only focused on image pre-processing (such as object detection or segmentation) while the rest of the works are concentrated on simple machine learning classification approaches due to hardware limitations. Nevertheless, the accuracy of the prediction task is of utmost importance but none of these solutions have focused on exploring the possibility of deploying most advanced and latest cutting-edge deep neural networks on mobile devices to improve the accuracy in both data pre-processing and classification tasks. On the other hand, it is also important for all these classifiers to be efficient in terms of mobile resource utilization. Thus, in this paper we believe, it is the high time to look into the competence of modern hardware and the state-of-the-art DNNs toward achieving robust, accurate and efficient on-device predictions.

	\vspace*{-3mm}
\section{Methods of Mobile Image Analysis, Data Acquisition and Pre-processing for Smart Health}
In this study, our foremost focus is on three different image-based deep learning powered on-device smart health diagnosis applications: 1) detection of eardrum abnormalities such as Otitis media, 2) detection of different nail fungus diseases such as Melanonychia and Naildystrophy, and 3) detection of skin lesion diseases such as Melanoma, Melanocytic nevus, Basal cell carcinoma and so on. In reality, smartphone captured images in visible light are generally not up to par with the images acquired under controlled clinical conditions. Thus, in this section we discuss different levels of image analysis and data pre-processing techniques that can be utilized in various application scenarios within on-device deep learning paradigm.

\begin{figure}[ht]
	\centering
	\includegraphics[width=0.8\columnwidth]{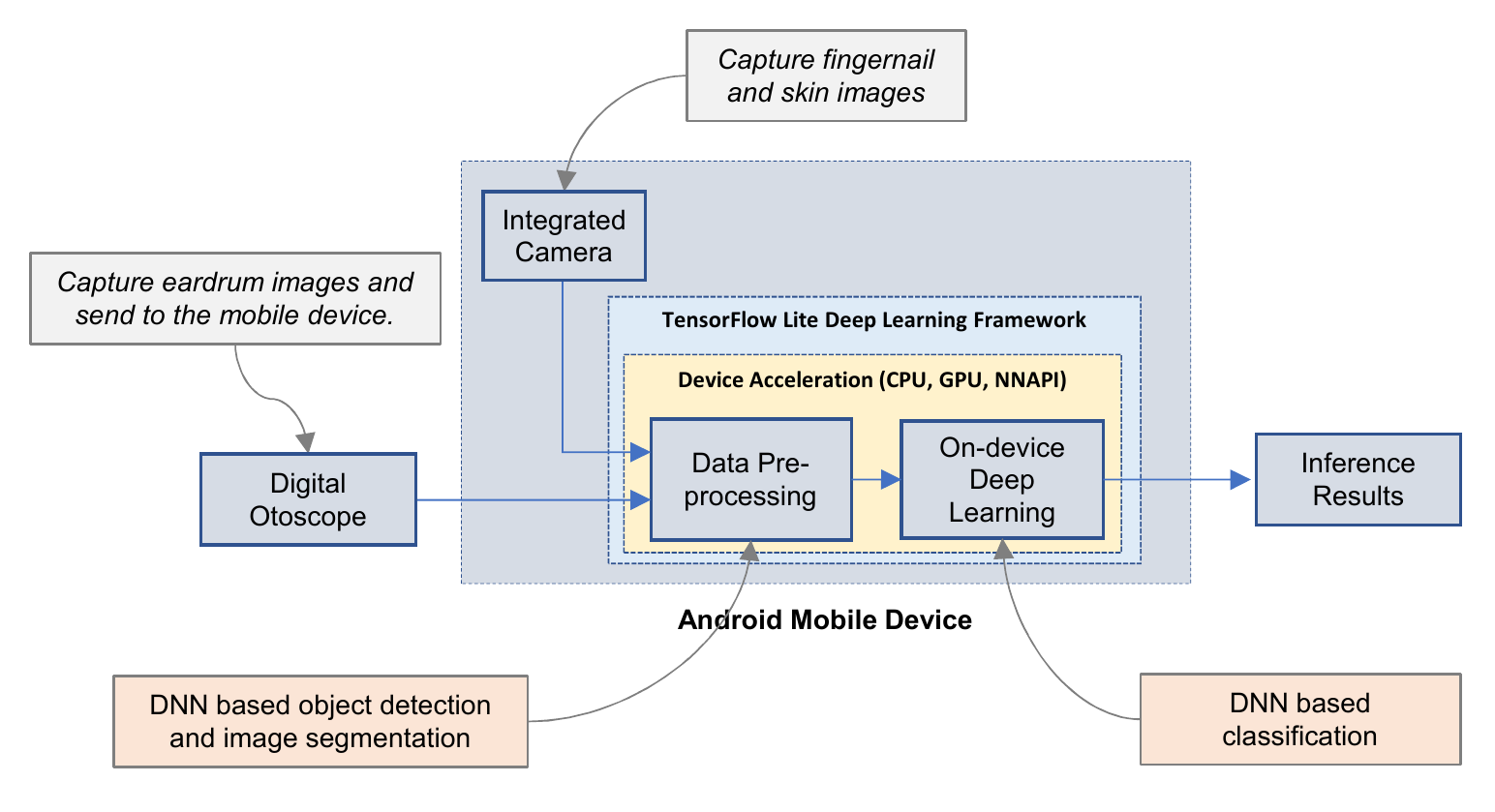}
	\caption{High level Block Diagram of Data Pre-processing and Disease Diagnosis.}
	\label{fig:block_diagram}
\end{figure}

\vspace*{-3mm}
\subsection{Data Acquisition and Pre-processing for Eardrum Abnormality Detection}
Abnormalities in eardrum, in particular, diseases such as Otitis media is a type of middle ear infection and one of the most common pediatric diseases usually develops as a complication of the upper respiratory tract \cite{bacsaran2020convolutional}. It usually has serious impacts which could even lead to undesirable conditions such as hearing losses and cognitive disorders. These abnormalities can be identified by examining the status of the eardrum, and in clinical practice, otoscope devices are frequently used to diagnose the eardrum. With the evolution of digital imaging technologies, today's latest digital otoscopes are capable of capturing high quality images of eardrum and by interfacing the device with a computer (or a smartphone) those images can be extracted for further observations. In our framework as depicted in Fig. \ref{fig:block_diagram}, we employed one of the latest digital otoscopes available in the market (Firefly DE570 Wireless Otoscope \cite{fireflyOtoscope}) and integrated it with the mobile application through an API such that images captured by the otoscope can be directly processed by on-device deep learning models.

\vspace*{-3mm}
\subsection{Data Acquisition and Pre-processing for Fingernail Abnormality and Skin Lesion Detection}
Nail fungus is another common type of disease which usually requires medical expertise to diagnose. However, with a proper trained deep learning model, different important features of nail fungus can be effectively identified. In our framework, we utilized the embedded camera of smartphone to capture the hand image and then use object detection algorithms to extract the nail parts of the fingers for further processing. In the case of skin lesion, again we utilized the embedded camera of smartphone to capture the skin image and then devised image segmentation techniques to extract the region of interest as depicted in Fig. \ref{fig:object_detection}.

\vspace*{-3mm}
\subsection{Effect of Object Detection and Image Segmentation for the Diagnosis Task}
\begin{figure}[ht]
	\centering
	\includegraphics[width=0.80\columnwidth]{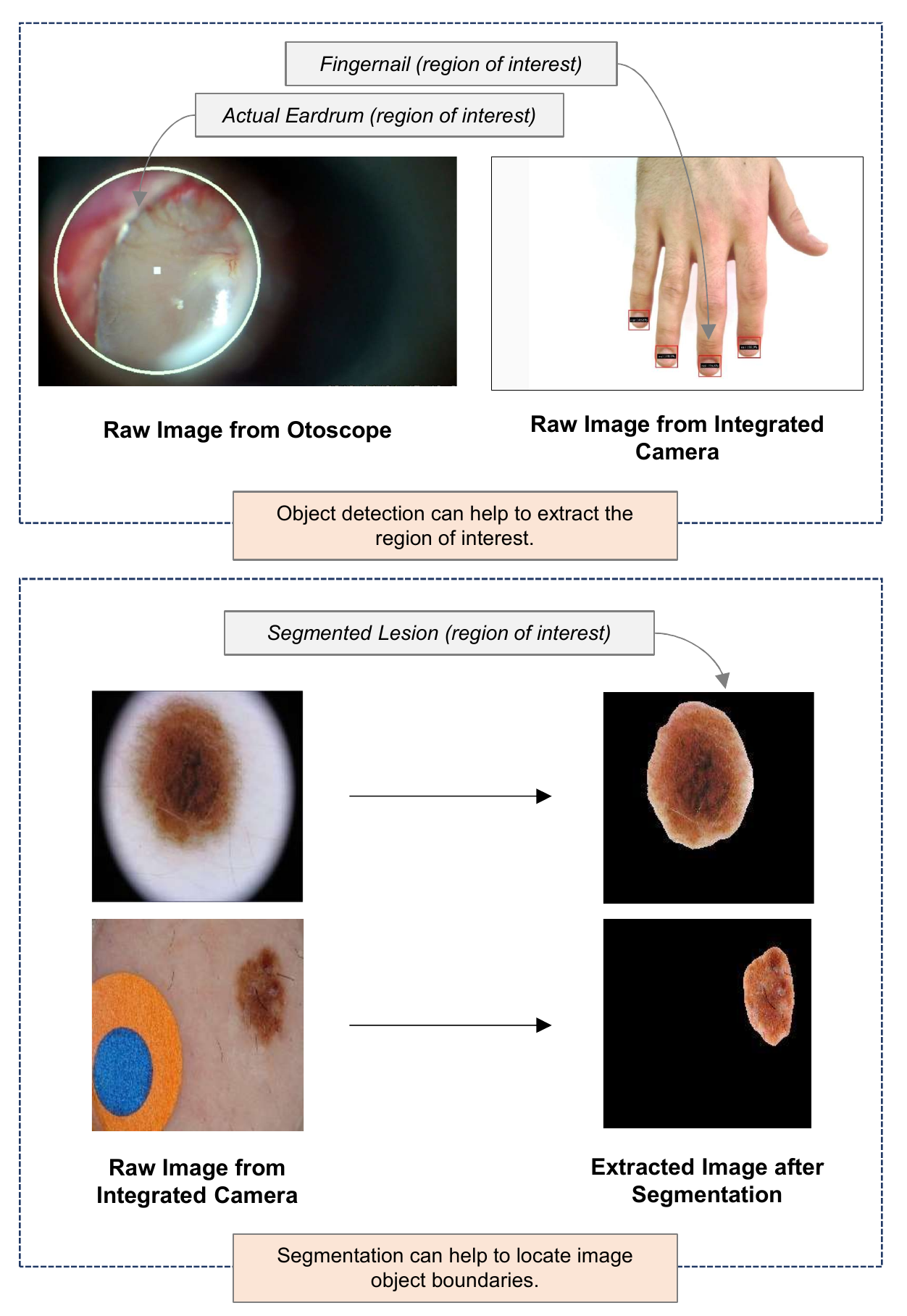}
	\caption{Use of Object Detection and Image Segmentation in the Data Pre-processing.}
	\label{fig:object_detection}
\end{figure}

Even though the image datasets that we used to train the machine learning models are originated from controlled clinical conditions with good quality, it is quite challenging to capture images with the same good quality by using a smartphone camera or otoscope due to various reasons such as different lighting conditions, angle of the capturing image, zooming factor of the lens, coverage of region of interest and so on. But, with an additional stage of image pre-processing, these complications can be minimized. In our solution, we utilized CNN-based object detection algorithms \cite{liu2016ssd} within the data pre-processing architecture to achieve a couple of objectives. For example, in the case of eardrum diagnosis, whenever the image is captured from otoscope the actual eardrum (region of interest) is not always positioned at the center of the image and its is neither fixed into a specific location. Thus, by using a well trained object detection algorithm the region of interest can be easily captured. Similarly, in the case of fingernail diagnosis, we employ object detection algorithms to extract the fingernails from an image of person's hand as shown in Fig. \ref{fig:object_detection}.

Meanwhile for the skin lesion diagnosis application, image segmentation serves as a pre-processing mechanism to extract the skin lesion part (region of interest) out from the images. In practice, smartphone based photographic images of skin sometimes can be disturbed/distorted with non-skin parts such as bandages and other backgrounds. In such circumstances, image segmentation greatly helps to detect the skin and extract the lesion out of the image for classification. Therefore, our solution is comprised with CNN-based latest image segmentation algorithms \cite{ibtehaz2020multiresunet} in terms of identifying the region of interest and improving the overall classification accuracy.

	\begin{figure*}[ht]
	\centering
	\includegraphics[width=0.80\textwidth]{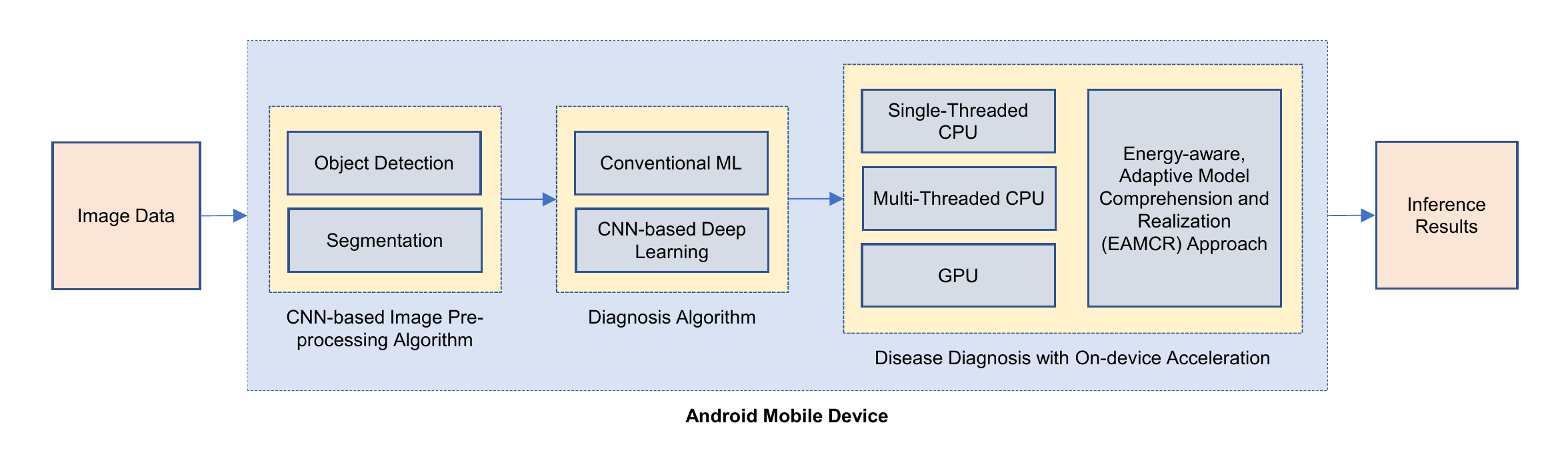}
	\caption{Overview of the Proposed Mobile Deep Learning Framework Implementation on Android Mobile.}
	\label{fig:implementation}
\end{figure*}

\vspace*{-3mm}
\section{Mobile Deep Learning Framework Implementation}
Based on the requirements captured for mobile based smart health applications and the solutions proposed in the recent literature toward efficiently utilizing deep learning models in mobile devices, we designed a framework for on-device deep learning and implemented a medical image diagnosis application for Android devices. Our solution consists of two main modules: 1) state-of-the-art data pre-processing module using CNN-based object detection and image segmentation algorithms, 2) disease diagnosis module using latest powerful deep learning image classification algorithms. At a high level,  as depicted in Fig. \ref{fig:block_diagram}, our solution works as follows. The Android application first takes the input from either internal camera or externally connected otoscope device, then process the image using either object detection or segmentation algorithm to extract the most important features (depending on the case) and finally, classify the image using cutting-edge deep learning models. Our solution is built on top of TensorFlow Lite \cite{tensorflowlite} open source deep learning framework and we implemented the necessary features/libraries to run the inference task of all the models on various hardware accelerators such as CPU, GPU and NNAPI on the mobile device. 

It is noteworthy to mention that, most of the implementations discussed in the literature are mostly based on simple (or mobile optimized) CNN architectures such as MobileNet \cite{howard2017mobilenets} or SqueezeNet \cite{iandola2016squeezenet} however none of them have implemented or explored the possibility of implementing recent state-of-the-art CNN architectures. In contrast, we have explored almost all the latest and state-of-the-art CNNs including Inception \cite{szegedy2016rethinking}, Xception \cite{chollet2017xception}, InceptionResNet \cite{szegedy2017inception} and EfficientNet \cite{tan2019efficientnet} on mobile devices for energy-aware smart health applications. 

\vspace*{-3mm}
\subsection{\textbf{Datasets and the Model Training}}
We trained a list of CNN models for each application scenario based on different image datasets using TensorFlow ML platform. For the eardrum scenario, we obtained a publicly available eardrum dataset \cite{bacsaran2020convolutional} which consists of 282 otoscope images in 500x500 resolution. It consists of both normal and abnormal images in 6 different classes. In addition, we used another approximately 200 publicly available eardrum images from Google. In the case of nail fungus diagnosis, we obtained 53k labeled image dataset \cite{modelonychomycosis} which consists of images among 6 different classes. Further, we employed the latest ISIC 2019 dataset \cite{isic2019} and 2017 dataset \cite{isic2017} for the skin lesion scenario. This dataset contains around 27k dermoscopic images across nine different diagnostic categories. For the eardrum and fingernail applications, we defined the classification task as binary classification: normal and abnormal while skin lesion is defined to be multi-class classification. It is important to mention that all of these dermoscopic images have been obtained under controlled clinical conditions; hence most of them are high quality images. However, our solution is based on photographic images captured using smartphone camera (or otoscope) and hence variability in factors such as lighting condition, angle and region of interest might affect the accuracy. Therefore, apart from the image pre-processing with object detection and segmentation approaches, we also employed different data augmentation techniques such as rotation and cropping to improve the model training.

\vspace*{-3mm}
\subsection{\textbf{Implementation on Mobile}}
As stated before, our implementation is mainly architectured on top of the latest TensorFlow Lite deep learning framework (Fig. \ref{fig:block_diagram}). The data pre-processing module of the solution consists of several CNN-based object detection and segmentation algorithms. Based on the recent advancements mentioned in the literature, we first experimentally evaluated the latest object detection algorithms including Faster RCNN \cite{ren2015faster}, Single Shot MultiBox Detector (SSD) \cite{liu2016ssd} and Efficientdet \cite{tan2020efficientdet} with regard to accuracy of prediction of the different object detection tasks in eardrum and fingernail diagnosis. In terms of training the object detection algorithm for fingernail diagnosis, we utilized a publicly available hand image dataset \cite{HandDataset} of 1218 samples. Then these samples were manually annotated using LabelImg image annotation tool \cite{LabelImg}. Once the annotation files are generated, 1000 out of those samples were used to train the object detection models and tested the accuracy as the number of detected objects (nails). Then, the best performing trained models were converted to TensorFlow Lite and deployed on the Android mobile. It is noteworthy that due to the complexity of the underlying CNN architecture, conversion of the Faster RCNN model (trained on fingernail dataset) was not a success. Hence, we utilized the SSD and Efficientdet based approaches when deploying on the mobile device. 

\begin{table}[]	
	\centering
	\caption{A Summary of CNN-architectures Deployed on Data Pre-processing Module for Object Detection and Segmentation.}	
	\includegraphics[width=0.8\linewidth]{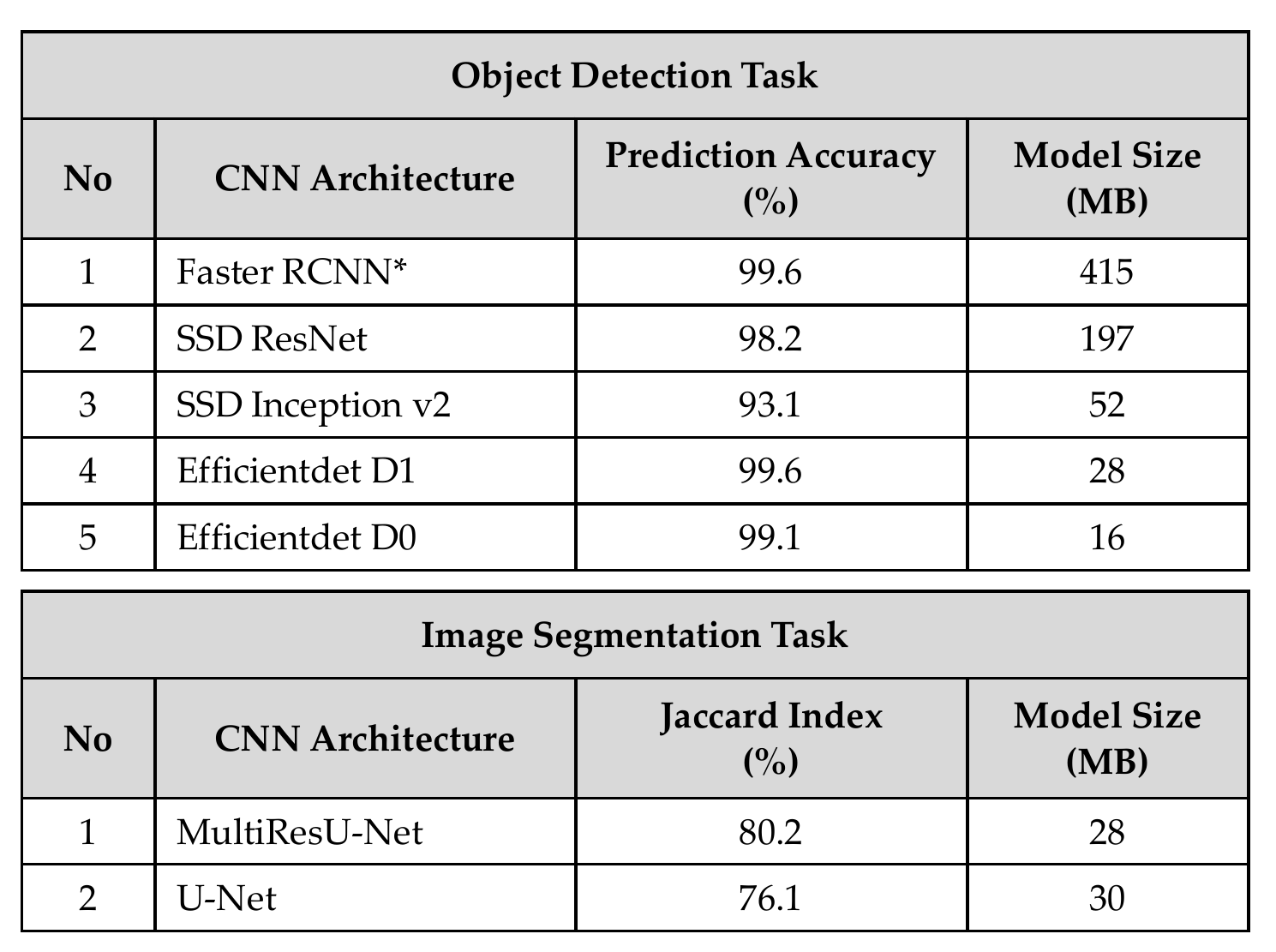}
	
\end{table}

Similarly, in terms of image segmentation, U-Net \cite{ronneberger2015u} and the recently proposed MultiResU-Net \cite{ibtehaz2020multiresunet} biomedical image segmentation algorithms have been implemented and evaluated in terms of model accuracy for the segmentation task in skin lesion diagnosis. In terms of training the segmentation model, we utilized the 2000 samples with masks in the ISIC 2017 dataset \cite{isic2017}. It is noteworthy that these 2000 samples were also included in the ISIC 2019 dataset. Therefore, during the testing phase we removed those repetitive samples from the ISIC 2019 dataset before conducting the experimental evaluations. Based on the results, MultiResUNet is deployed on mobile as the data pre-processing algorithm for skin lesion diagnosis. To measure the results of segmentation, the Jaccard index \cite{mcguinness2010comparative} is used as follows,

\[J(X,Y) = \frac{|X\cup Y|}{|X\cap Y|} \]

\noindent
where $X$ represents the ground truth binary segmentation mask while $Y$ is the predicted binary segmentation mask. A summary of CNN-based architectures devised in data pre-processing module is shown in Table 1 along with the prediction accuracy and the model size.

Once the image is pre-processed, it is then fed into the diagnosis module for the classification/disease diagnosis task. For that, we devised two different approaches through both conventional machine learning and CNN-based deep learning methods.

\subsubsection*{\textbf{Conventional Machine Learning vs CNN-based Deep Learning for the Classification Task}}

While CNN-based image classification algorithms provide significant benefit in terms of accuracy, it is equally important to ensure the energy efficiency of the application when it is deployed on resource constrained environments. Thus, with regard to disease diagnosis, our solution consists of both CNN-based architectures and conventional machine learning approaches. We have empirically evaluated both these architectures toward finding an efficient on-device ML model for each smart health application scenario. 

In the case of conventional ML, during the training we employed OpenCV library \cite{bradski2008learning} to extract the reliable features from images (feature extraction). Thereafter, several different classifiers including K Nearest Neighbor (KNN), Support Vector Machine (SVM), Logistic Regression, Random Forest and Gradient Boosting have been trained in Python in terms of finding the best classifier for each scenario/task. Once the training is completed, we picked the best performing classifiers to implement them on Android for the diagnosis task. It is noteworthy that the gradient boosting approach outperforms rest of the classifiers significantly in all cases. In fact, for the implementation of gradient boosting on Android, we utilized both LightGBM \cite{ke2017lightgbm} and XGboost \cite{chen2016xgboost} approaches which are among the top of efficient and scalable implementation of gradient boosting. Our approach of utilizing conventional ML in disease diagnosis on skin lesion is summarized in Table 2 including prediction accuracy of the model, size of the model and inference time on different accelerators.

\begin{table}[]	
	\centering
	\caption{A Summary of Conventional ML Deployed on Mobile Device in terms of Skin Lesion Diagnosis Task.}	
	\includegraphics[width=0.8\linewidth]{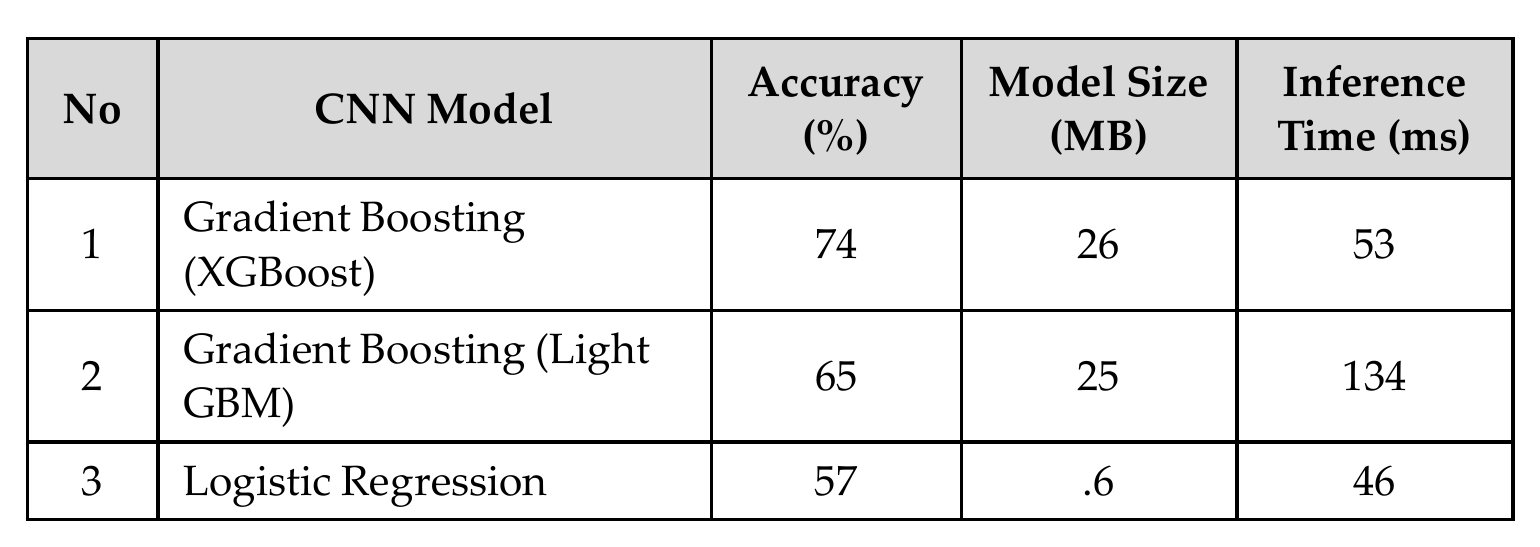}	
\end{table}

\begin{table}[]	
	\centering
	\caption{A Summary of CNN-based ML Architectures deployed for On-device Deep Inference.}	
	\includegraphics[width=0.8\linewidth]{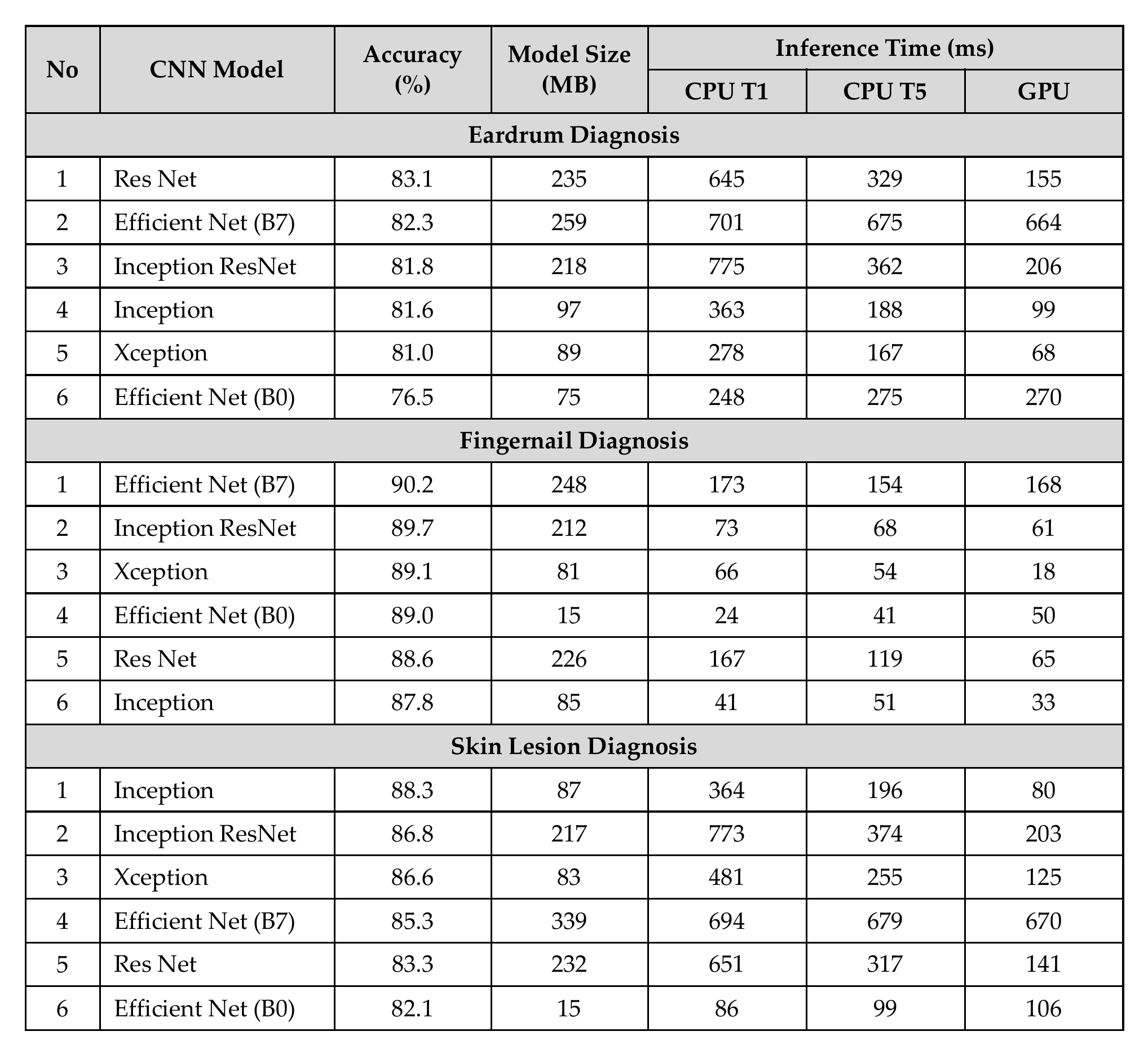}
	
\end{table}

Comparably, for the CNN-based approach, a set of state-of-the-art deep learning models have been implemented in Python and trained on TensorFlow. Thereafter, these trained models have been converted to TensorFlow Lite and then deployed on the mobile device for the inference results. A summary of CNN architectures deployed on mobile is listed in Table 3. In addition to the inference task, we also included a relevant set of additional line of codes to capture the model performance during the runtime including inference time, CPU/GPU utilization, memory usage and so on. Moreover, in order to capture the energy consumption, we utilized the Battery Manager class in Android along with our implementation of battery status API to capture battery usage/drain (in mA), total capacity of the battery (in mAh) and the voltage (in V) of each sub task in the smart health application.

\vspace*{-3mm}
\subsection{Calculating the Energy Efficiency of Algorithms through Deep Learning Efficiency Index (DLEI)}
\begin{figure}[ht]
	\centering
	\includegraphics[width=0.95\columnwidth]{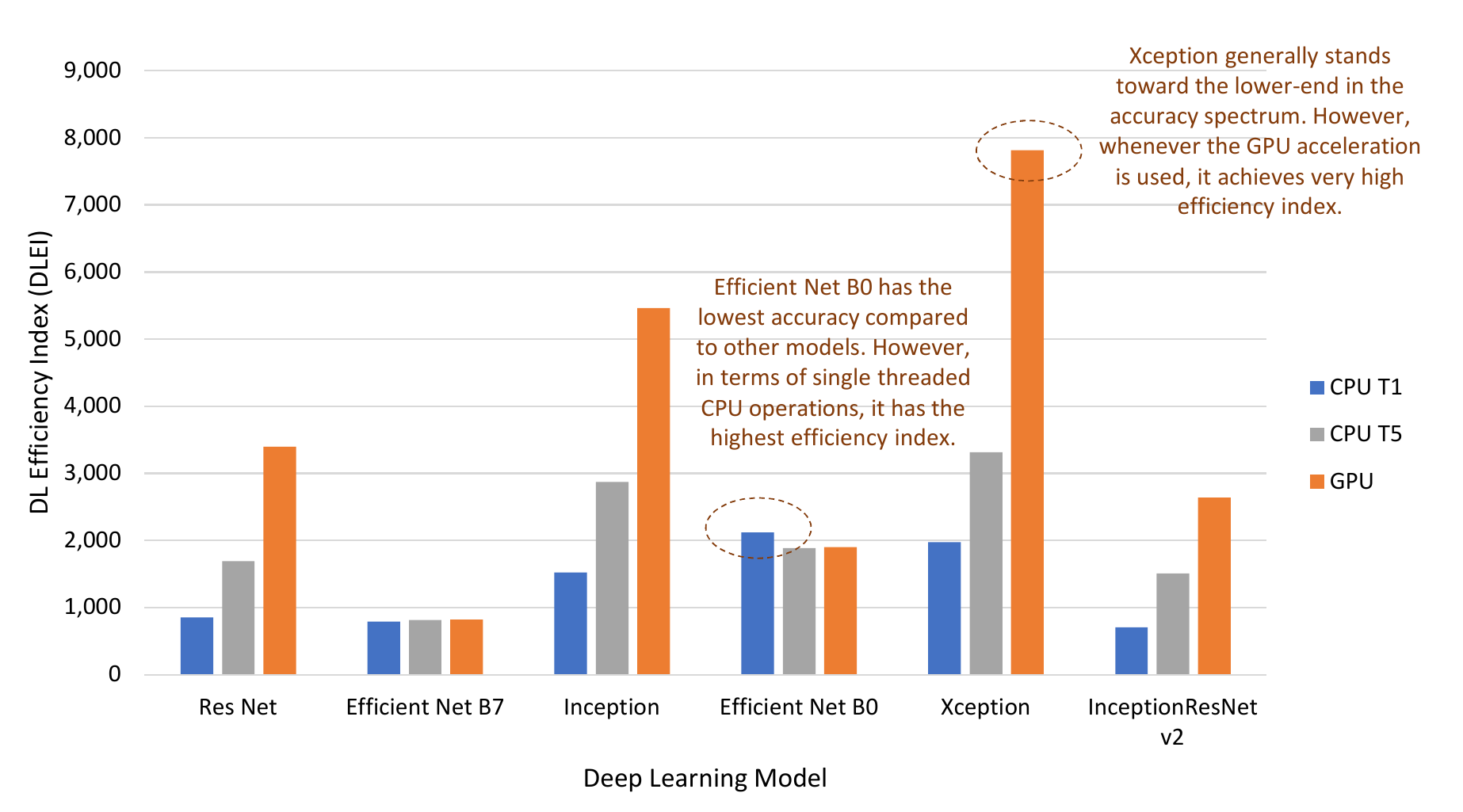}
	\caption{Comparison of Deep Learning Efficiency Index Values for Different Accelerators (Eardrum Diagnosis).}
	\label{fig:efficiency_index}
\end{figure}

It is well understood that some deep learning models consume more energy to produce high accurate inference whereas some models consume less energy and still produce inference results with reasonable accuracy (mostly due to the complexity of the model). Hence, in practice, it is important to analyze energy characteristics of these deep learning models to determine the efficiency in terms of energy consumption. Thus, we devised on-device Deep Learning Efficiency Index (DLEI) as an indicator to compute the performance of energy consumption in an on-device machine learning task. It can be seen as a key performance indicator (KPI) with the goal of reducing energy demand while improving the overall accuracy of the inference task. DLEI is computed as follows,

\[Deep Learning Efficiency Index = \frac{A_{t}}{AE_d} \]

where $AE_d$ is the actual energy demand/consumption for a given inference task and $A_{t}$ is the accuracy of the inference task. A summary of how DLEI varies across different DL models when deployed on various accelerators for the eardrum analysis is shown in Fig. \ref{fig:efficiency_index}. As it explains, when the model accuracy is higher it contributes to improve the index whereas when the energy consumption is higher, it contributes to decrease the index value. In a nutshell, the higher the efficiency index is, better the model to be used in practice as it requires low energy to produce better results. 

As the results shown in the Fig. \ref{fig:efficiency_index} advocate, in general, performing the inference task with multi-threaded CPU can produce higher efficiency index than single-threaded CPU operations. On the other hand, changing the accelerator to GPU, yields better efficiency index compared to CPU operations. However, it is noteworthy that EfficientNet models perform slightly different where there is no much of a significant difference by running the inference task on CPU or GPU. Nevertheless, EfficientNet B0 has the lowest accuracy compared to other models in the eardrum analysis task. But, in terms of single-threaded CPU operations, it has the highest efficiency index meaning when the application is running on single-threaded CPU, EfficientNet B0 is the most efficient model to be used in practice as it withstands longer operation time. 

For the eardrum analysis task, ResNet produces the highest accuracy (refer Fig. \ref{fig:eardrum_modelsize}). However, whenever the accelerator is switched back to multi-threaded CPU or GPU, Xception has the most prominent efficiency index making it most reliable model to be used in practice in terms of energy efficiency. Conclusively, we observed a similar tendency in results for both fingernail and skin lesion analysis tasks; hence, energy efficiency of DL task is one of the vital factors stands alongside with accuracy and inference time to be carefully ascertained by the on-device deep learning applications in practical deployment scenarios. 

\vspace*{-3mm}
\subsection{On-device Deep Learning with Energy-aware, Adaptive Model Comprehension and Realization (EAMCR) Approach for Resource Contested Environments}
As mentioned before, whenever the deep inference is performed locally, managing the balance in between energy consumption and the inference time is the key governing factor for any mobile application. Based on our experimental observations, and different CNN and conventional ML model parameters, we identified an initial framework to be deployed for on-device deep learning particularly in smart health applications through an adaptive model comprehension and realization approach. Over the past, different approaches have been proposed in the literature such as MCDNN \cite{han2016mcdnn} to address this problem using new model optimization/scheduling techniques for cloud-backed mobile devices. However, EAMCR takes a different yet simple approach. Our intention is to focus on how to utilize existing models in a way such that system could adapt to energy demands by switching to different variants of the existing models and accelerators, but still maintaining the same (or comparable) level of accuracy. The framework has an integrated decision engine with two operating modes. The first mode is the open access learning mode where application user has the freedom to choose any of the deployed CNN in any application scenario, regardless of the energy consumption. However, whenever the remaining battery capacity becomes a certain threshold $th$ (this value can be configured by the user), the application automatically switched to an energy efficient mode where inference task will be performed based on two governing factors. First, user has the option to select the intended accuracy region for the inference results. Once it is picked, then it is required to choose how many hours of operation that user is planning to operate the device. Once these values are obtained, the appropriate CNN model is automatically chosen by the decision engine to run the inference. Following simple energy calculation is employed during the time of deciding which model to be picked during the runtime.

\[Estimated Usage Time = \frac{Remaining Battery Capacity}{Total Discharge} \]

\begin{figure}[ht]
	\centering
	\includegraphics[width=0.9\columnwidth]{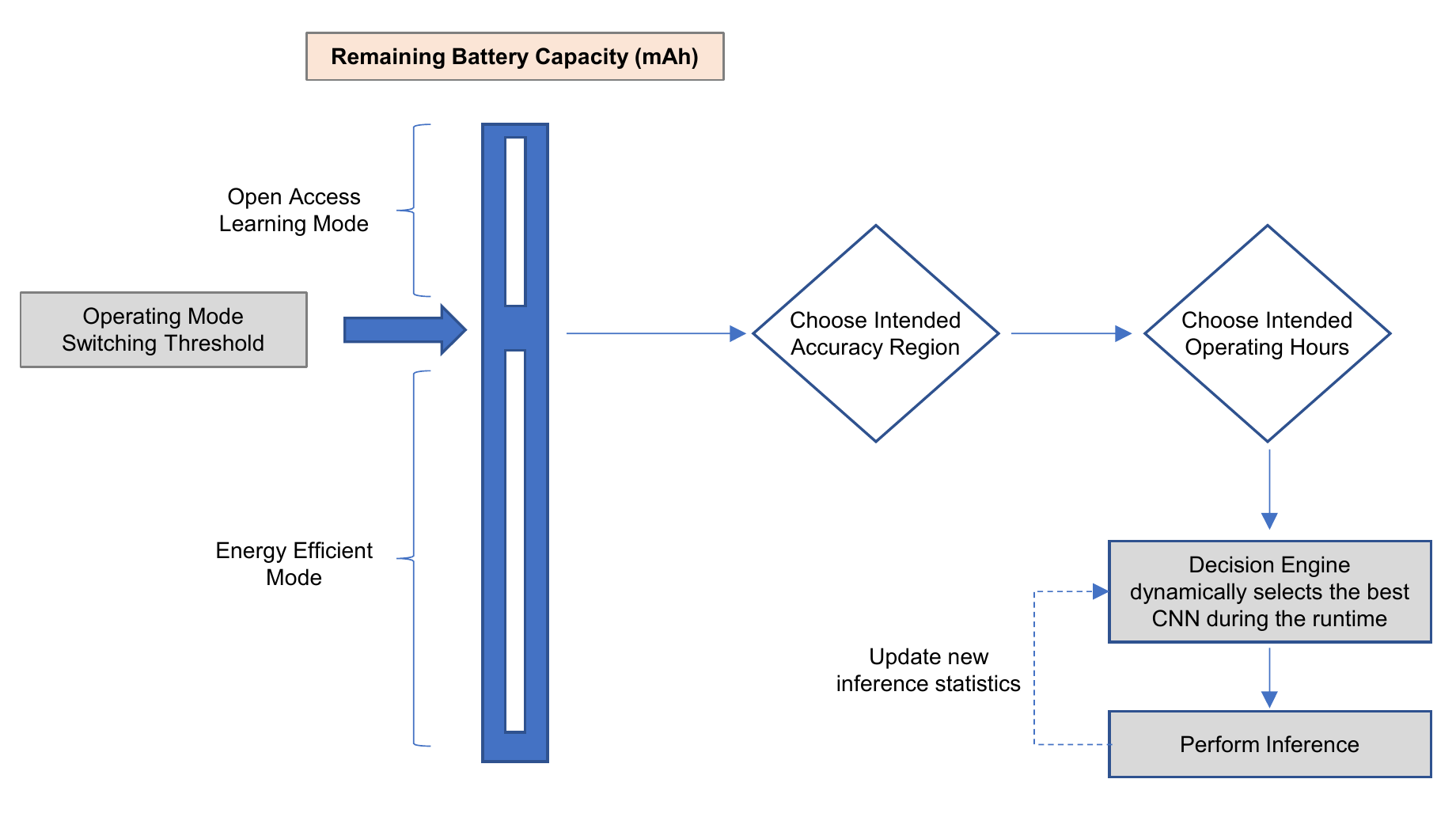}
	\caption{A High Level Overview of the Operations of Decision Engine During the Runtime in EAMCR Approach.}
	\label{fig:new framework}
\end{figure}

The remaining battery capacity (mAh) is extracted during the runtime using the integrated Battery Status API and the total discharge of current (mA) can be obtained per accelerator per CNN basis from the initial experiments performed. A high level overview of the initial framework is depicted in Fig. \ref{fig:new framework}. To adopt to the environment dynamically, decision engine needs to update the inference statistics periodically. Hence, after each inference task, the statistics are feedbacked to the decision engine to tuneup the model selection parameters.

	\vspace*{-3mm}
\section{Performance Evaluation}
Based on a set of empirical measurements obtained during the inference tasks of three smart health/telemedicine application scenarios with current state-of-the-art CNNs and conventional ML approaches, in this section we present evaluation results mainly in terms of inference time, impact of model size and energy efficiency.

\vspace*{-3mm}
\subsection{Experimental setup and comparison}
The inference experiments were mainly conducted using following three mobile devices: Google Pixel 4 XL, OnePlus 7, Samsung Galaxy S10e. All three devices are equipped with the same Qualcomm Snapgragon 855 Octa-core CPU and Adreno 640 GPU. We evaluated the performance of six different CNNs architectures along with our Android application in three application scenarios as summarized in Table 3. It is noteworthy to mention that each experiment has been repeated for ten times (for each mobile device) and the average values were reported. We instrumented different internal logging mechanisms within the Android application to report inference time, CPU and memory usage, and amount of battery usage. Moreover, before each experiment, we ensured the same level of battery capacity (100 percent) and the same set of background processes during the time of performing the experiment. For each experiment, application first randomly picks one of the pre-loaded testing images and thereafter user has the option to select the ML architecture and the accelerator (whether single-thread CPU, multi-thread CPU or GPU). Once it is selected, inference task is performed. This process is repeated for ten times and the average values are recorded.

\begin{figure}[ht]
	\centering
	\includegraphics[width=0.8\columnwidth]{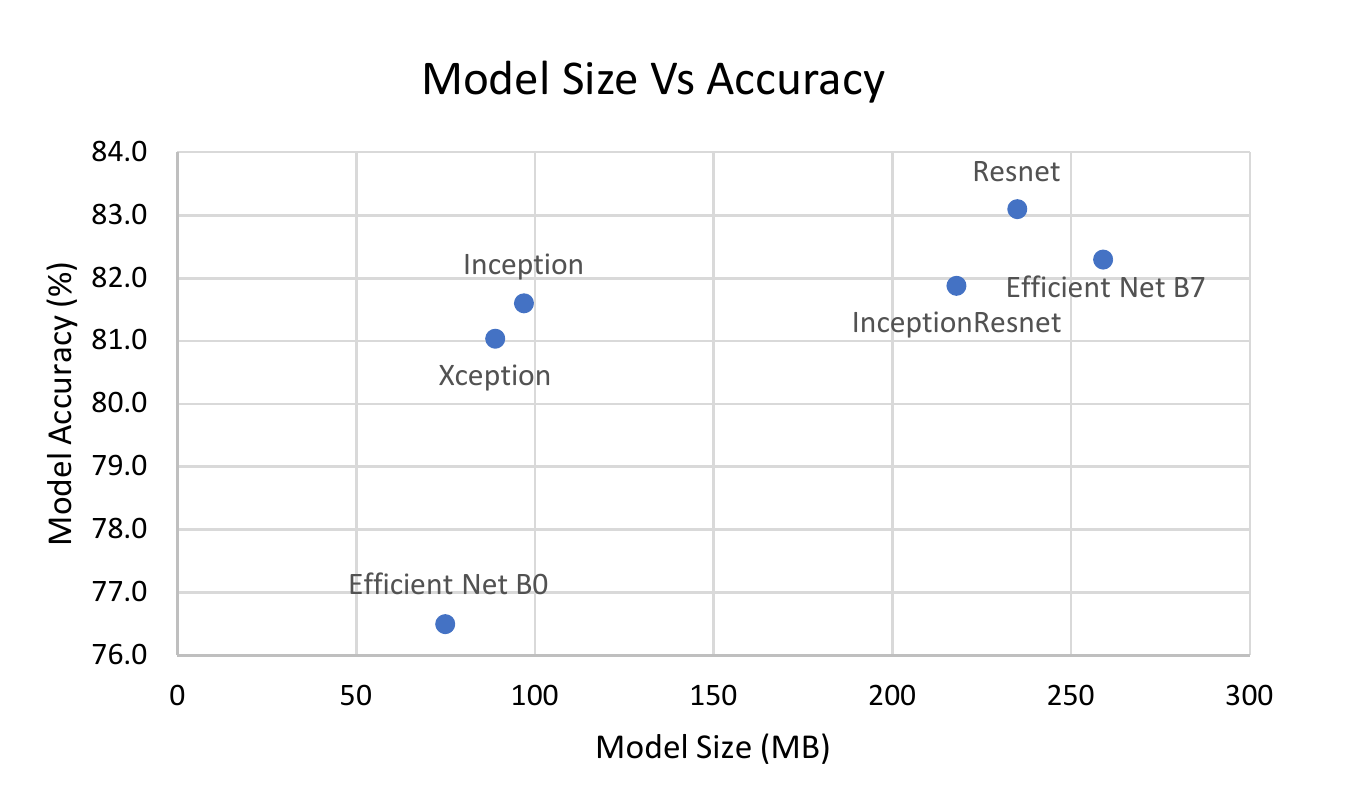}
	\caption{Comparison of Size of the Model against Accuracy for the Eardrum Diagnosis Task.}
	\label{fig:eardrum_modelsize}
\end{figure}

\vspace*{-3mm}
\subsection{On-device Performance Analysis}
Compression of deep learning models by limiting the number of layers is one of the known approaches to run deep inference on mobile. However, it has significant trade-off in terms of accuracy. In this section we analyze the significant parameters/metrics that govern the efficiency of on-device deep learning for the long run.

\subsubsection{Impact of model size}
First we looked into the impact of model size for the inference time. It is well understood that each ML architecture is unique where the number of layers and hyper-parameters are different, resulting diverse levels of accuracy and model sizes. Fig. \ref{fig:eardrum_modelsize} shows how the size of the model changes when the accuracy is increased. Throughout the experiments we observed that ResNet, EfficientNet (B7), and Inception ResNet always have comparatively larger model size while rest of the networks are smaller in size, for all three application scenarios. In a nutshell, whenever the number of parameters involved in the CNN is higher, the accuracy is getting better; however, on the flip side, model size is also getting larger.

\subsubsection{Impact of latency}
\begin{figure}[ht]
	\centering
	\includegraphics[width=0.8\columnwidth]{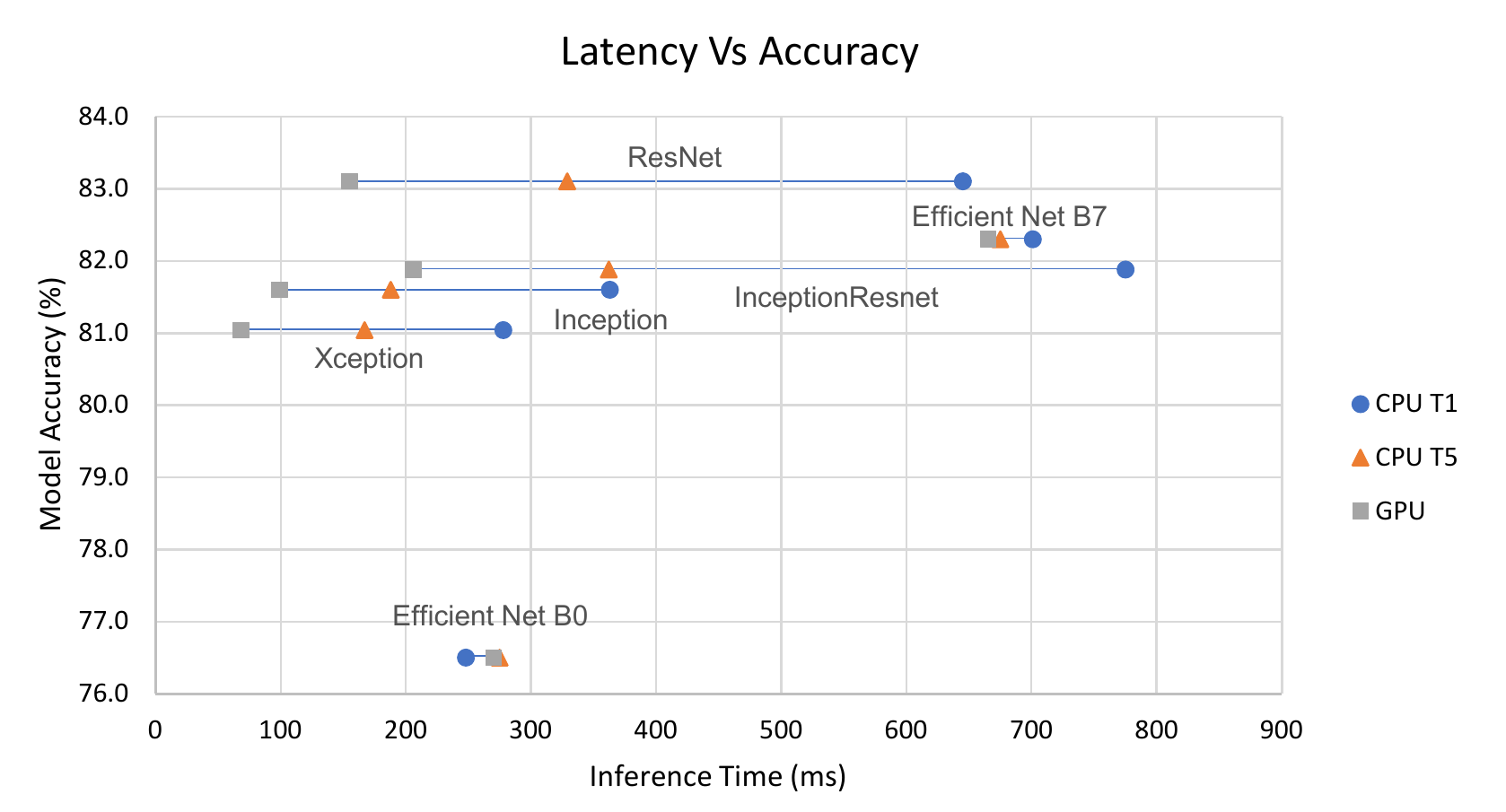}
	\caption{Comparison of Inference Time for Different Accelerators (Eardrum Diagnosis).}
	\label{fig:eardrum_latency}
\end{figure}

Next, we analyzed what is the impact for the inference time when the accuracy is increased. It is well understood that accuracy is one of the significant parameters, especially in healthcare applications. Hence, maintaining higher level of accuracy while managing the inference time within a substantial margin is a balancing act that has to be performed by the data driven intelligence. Therefore, we explored the possibility of achieving lower inference time while maintaining the same accuracy by utilizing different accelerators including hardware acceleration approaches. Fig. \ref{fig:eardrum_latency} shows the performing range of different accelerators including single-threaded CPU (CPU T1), multi-threaded CPU (CPU T5) and GPU. For example, while ResNet achieves the best accuracy for the inference task it's inference time whenever it is running on single-threaded CPU is much higher (645ms) compared to the other models. However, it was noticed that the same task can be performed on GPU with less amount of time (155ms) which is 4x faster than the aforementioned.

It was clearly observed that high accurate models are usually packaged in larger sizes hence suffer from high inference time. However, our intention with these comparison is to employ other means (such as additional accelerators) to reduce this inference time such that overall efficiency of the inference task is increased.

Now that the different accelerators are utilized, it is important to analyze the energy consumption of these accelerators in order to decide whether they are feasible enough to deploy in practice, especially in resource contested environments.

\subsubsection{Impact of Energy Consumption}
\begin{figure}[ht]
	\centering
	\includegraphics[width=0.9\columnwidth]{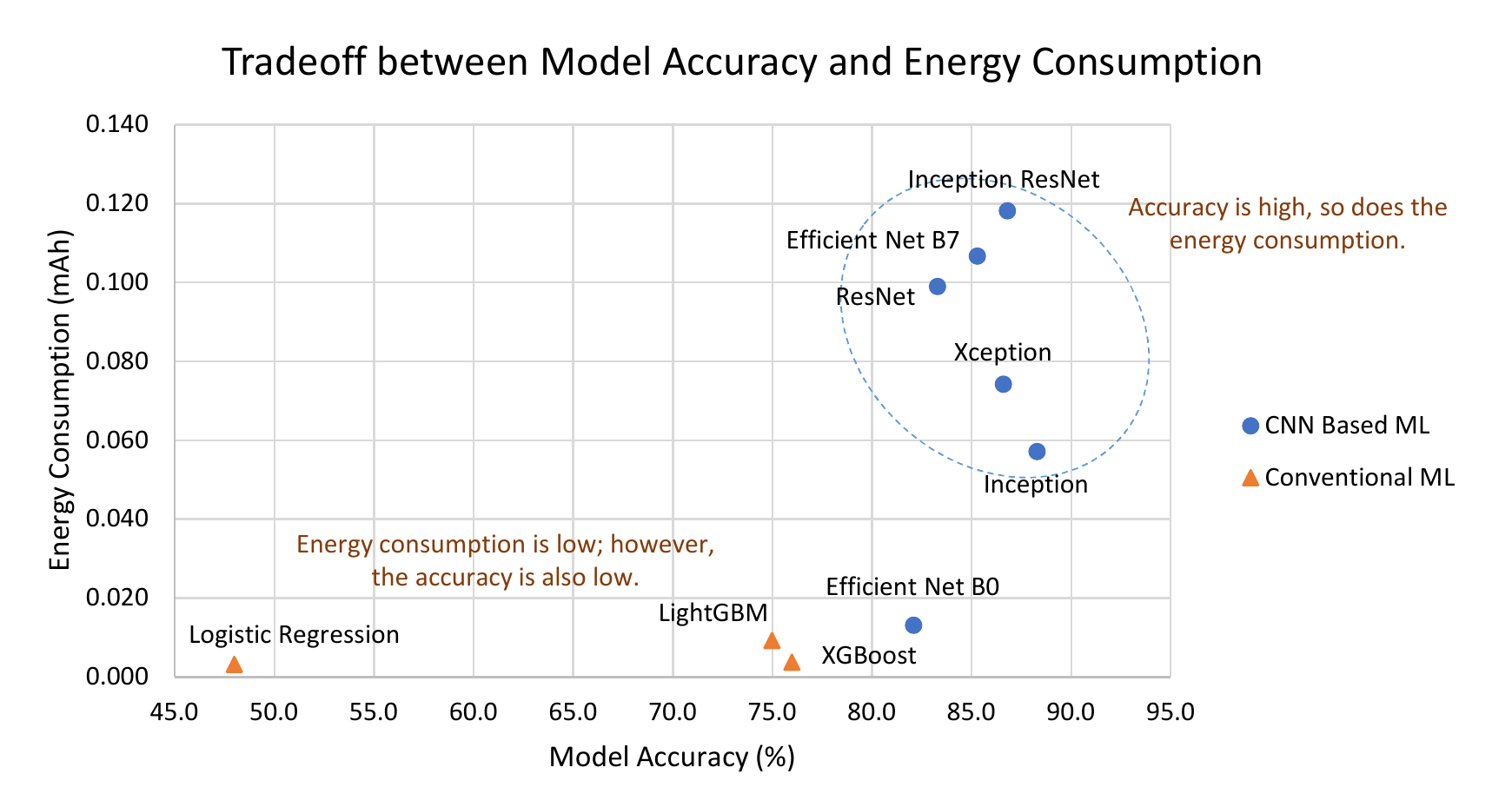}
	\caption{The trade-off between Accuracy of the Model and Energy Consumption.}
	\label{fig:tradeoff}
\end{figure}

We conducted a series of experiments to capture the energy consumption of both the conventional and CNN-based machine learning approaches. By utilizing the Battery Manger class in Android, along with our own implementation of battery status API, the battery voltage (in V) and the corresponding dissipating current (in mA) for each inference task have been recorded. In a nutshell, Fig. \ref{fig:tradeoff} shows the trade-off between model accuracy and the energy consumption of two of the on-device machine learning approaches. 

\begin{figure}[ht]
	\centering
	\includegraphics[width=0.8\columnwidth]{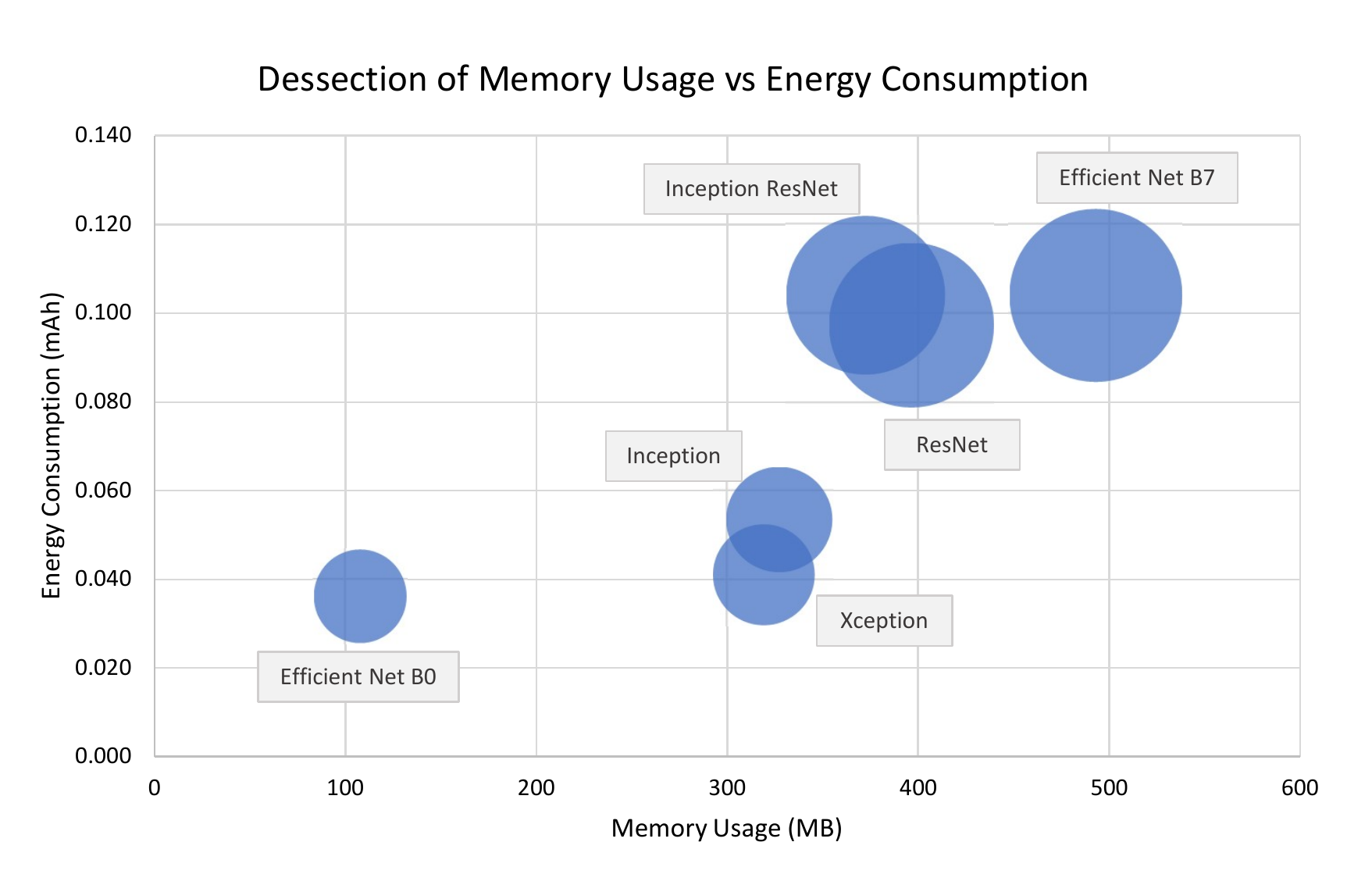}
	\caption{Higher the model size, (a) high consumption of memory resulting (b) high energy consumption. The buble size indicates the size of the model.}
	\label{fig:memory}
\end{figure}

It has been clearly identified that, in general, CNN-based models consume more resources while producing higher accurate results whereas conventional ML models consume only less computing resources with a considerable reduction in terms of accuracy of the results. Typically, running larger size CNNs require a lot of memory bandwidth to fetch all the model parameters (such as weights) and higher amount of computation to do dot products \cite{han2015deep}. Thus, as the summary of results depicted in Fig. \ref{fig:memory} describes, when the model size is larger it consumes more computing and memory resources, resulting higher level of energy consumption.

\begin{figure}[ht]
	\centering
	\includegraphics[width=0.8\columnwidth]{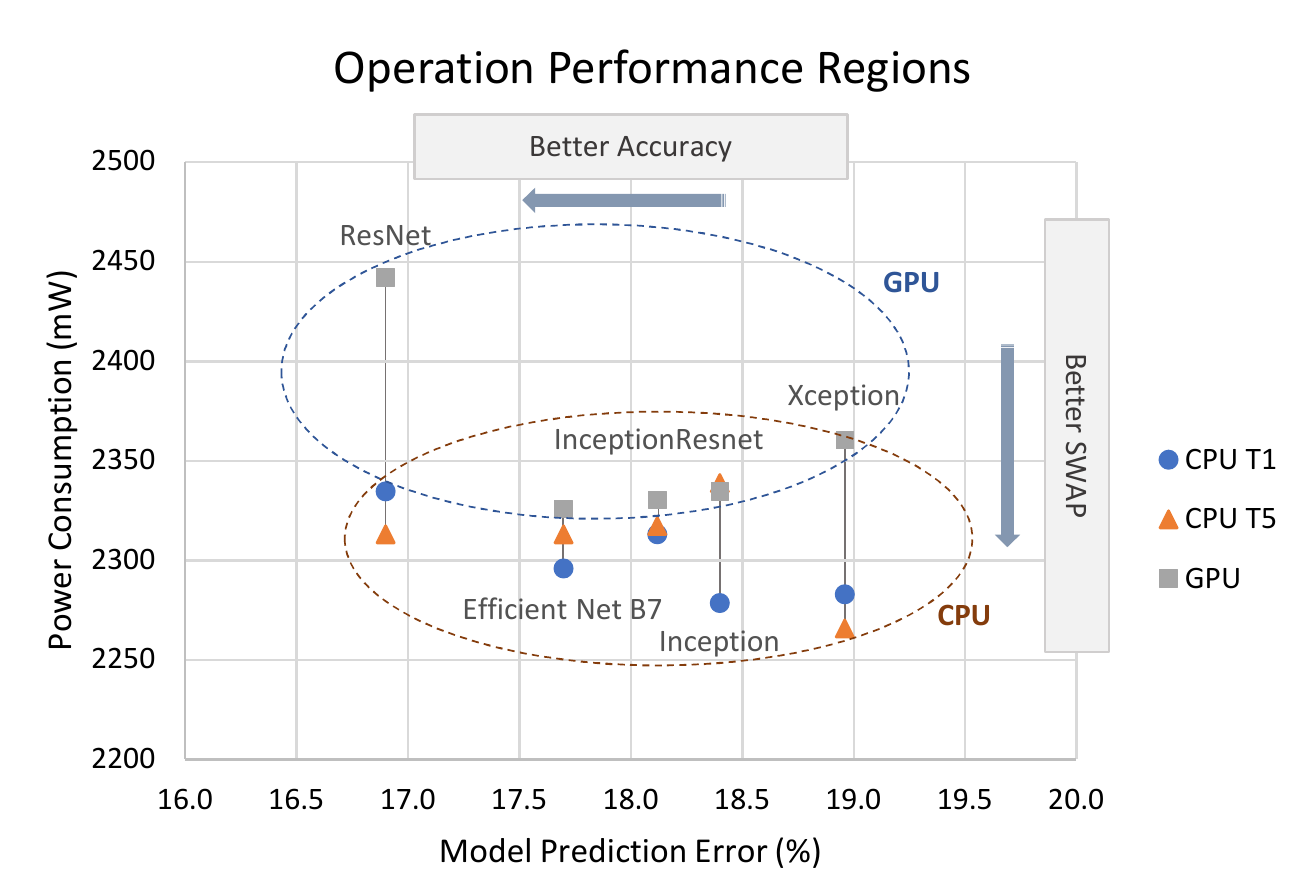}
	\caption{Operation Performance Regions of Different Accelerators for the CNN-based Implementations for Eardrum Diagnosis.}
	\label{fig:eardrum_operation_regions}
\end{figure}

Nevertheless, in healthcare applications, accuracy is of utmost importance so we further investigated the main contributing factors for high energy consumption in those CNN-based models. Accordingly, first we analyzed the power consumption of each CNN model against each different accelerator as depicted in Fig. \ref{fig:eardrum_operation_regions}. By utilizing the battery status API, the power consumption of each CNN model is tabulated against model prediction error. 

\begin{figure}[ht]
	\centering
	\includegraphics[width=0.9\columnwidth]{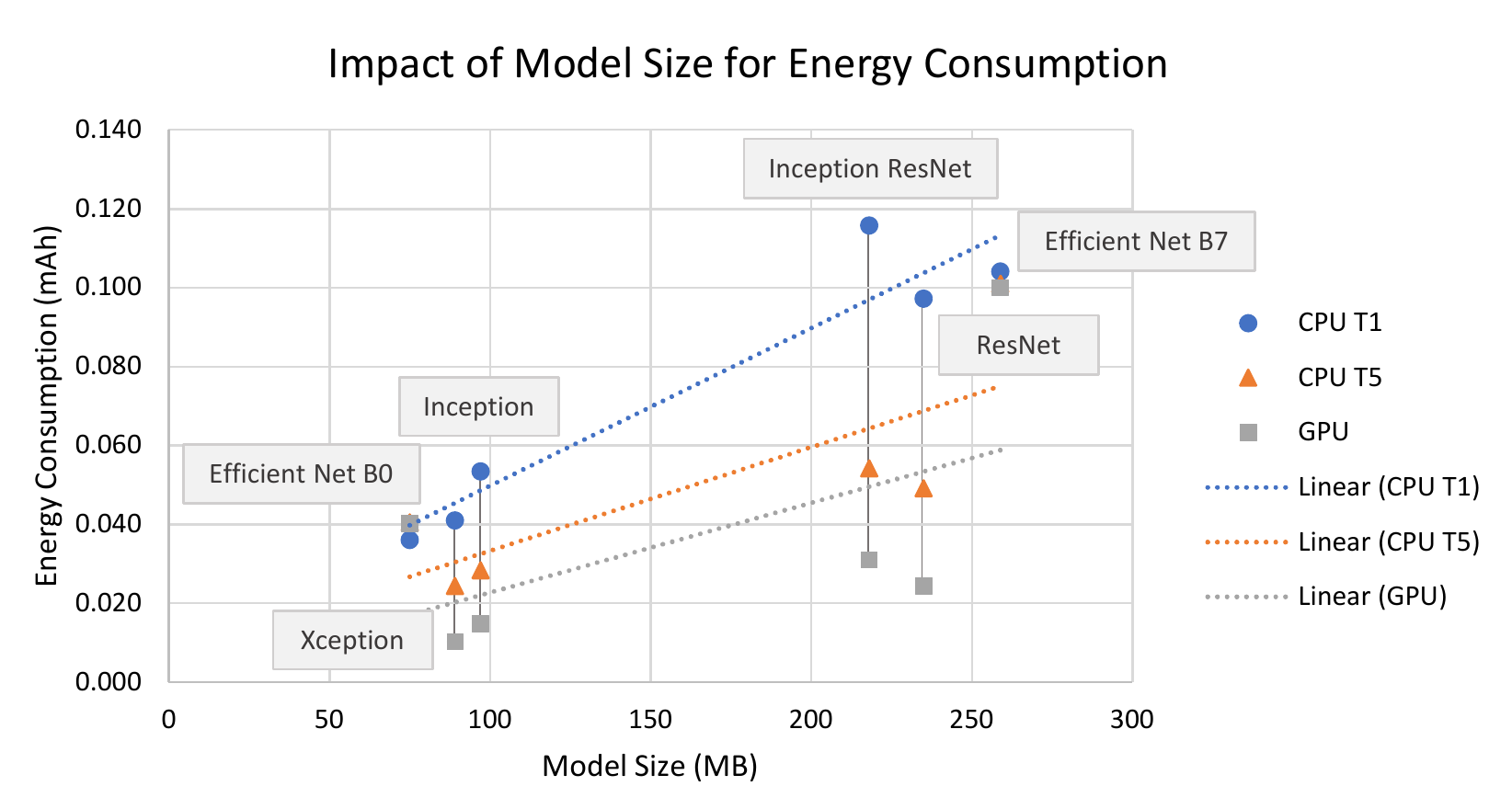}
	\caption{Comparison of Overall Energy Consumption on Different CNN Models (Eardrum Diagnosis).}
	\label{fig:eardrum_energy}
\end{figure}

It was observed that GPU is operating at the higher region by consuming more power (\texttildelow 2400mW for eardrum diagnosis) while CPU is operating in the lower region (\texttildelow 2290mW). However, it is noteworthy that since the GPU inference time is much shorter compared to the that of CPU, we observed a considerable reduction in overall energy consumption when the inference is performed on GPU. A summary of overall energy consumption for each inference task on different accelerators for the eardrum diagnosis is depicted in Fig. \ref{fig:eardrum_energy}. 

Similarly, we observed the same trend in energy consumption for the other two (fingernail and skin lesion) application scenarios as well.

\subsubsection{Impact of Energy-aware, Adaptive Model Comprehension and Realization (EAMCR) Approach}
\begin{figure}[ht]
	\centering
	\includegraphics[width=0.95\columnwidth]{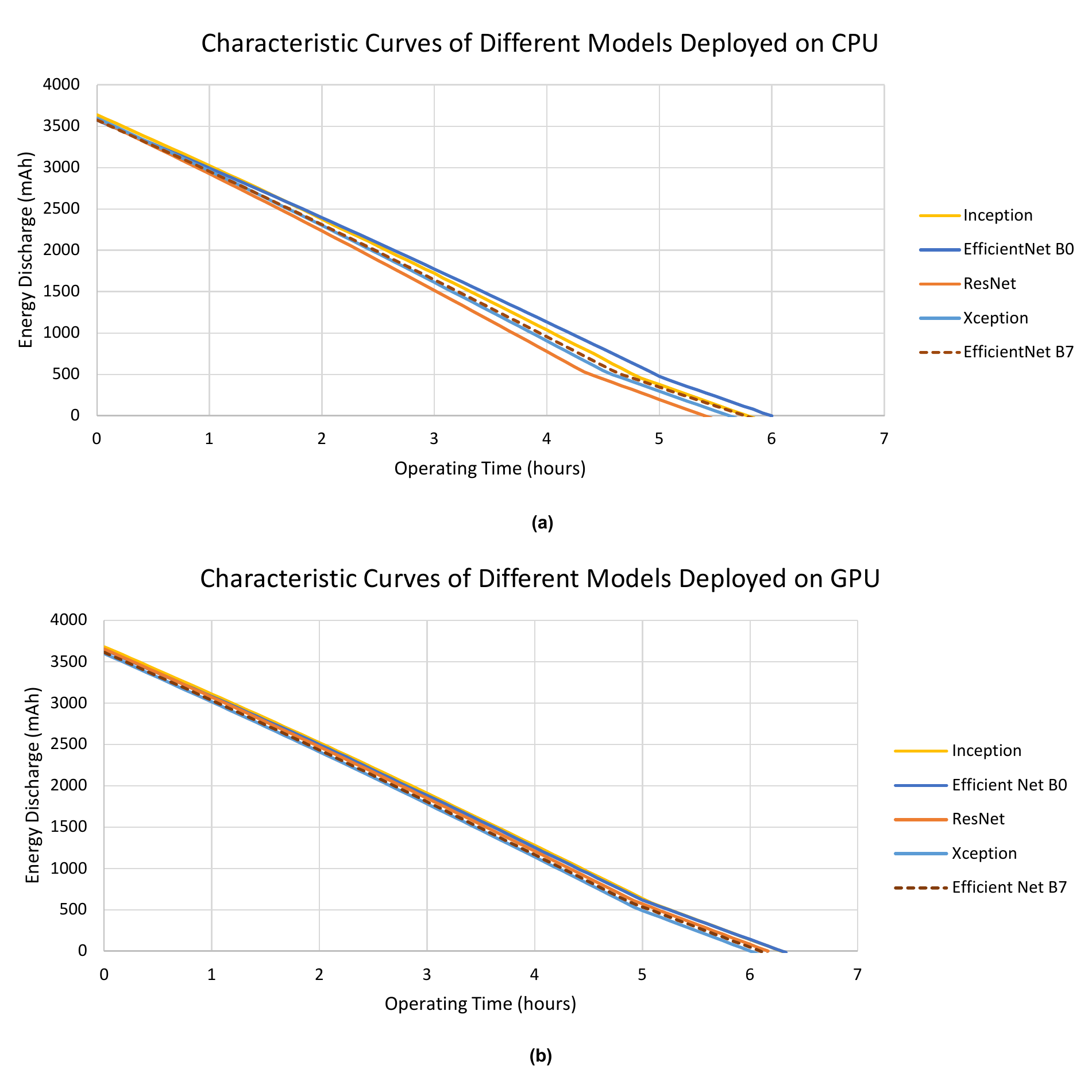}
	\caption{Energy Characteristics of Different Models when Deployed (a) on CPU (b) on GPU for Skin Lesion Diagnosis Task.}
	\label{fig:energy char}
\end{figure}

Based on the results of the initial findings, we implemented the EAMCR framework on Android mobile and empirically evaluated the performance along with a series of experiments on skin lesion diagnosis. First, the open access learning mode has been evaluated over the time. For these experiments, a OnePlus 7 Pro mobile device (Android 10 with design capacity of 4000mAh battery) has been utilized and we forcefully disabled all the background tasks of the mobile device such as WiFi, Mobile Data and Bluetooth, and we ensured the battery level of 100 percent at the start of each experiment. In addition, screen brightness level of 50 percent is set throughout the experiment. The energy characteristic curves for the different accelerators are shown in the Fig. \ref{fig:energy char}. It is understood that, in practice, use of other background tasks (for example web surfing with WiFi) might have higher impact on the overall energy demand, however at this point we leave that as a potential future work.

As the Fig. \ref{fig:energy char} describes, EfficientNet B0 outperforms all other models in terms of operating time when it is deployed on CPU whereas Inception (which has the highest accuracy) slightly outperforms all other models when the same task is deployed on GPU. Moreover, ResNet has the least operating time when deployed on CPU while Xception has the least operating time when the task is deployed on GPU. Thereby, on average we noticed at least \texttildelow30 minutes of difference in operating time when the accelerators are switched. 

\begin{figure}[ht]
	\centering
	\includegraphics[width=0.95\columnwidth]{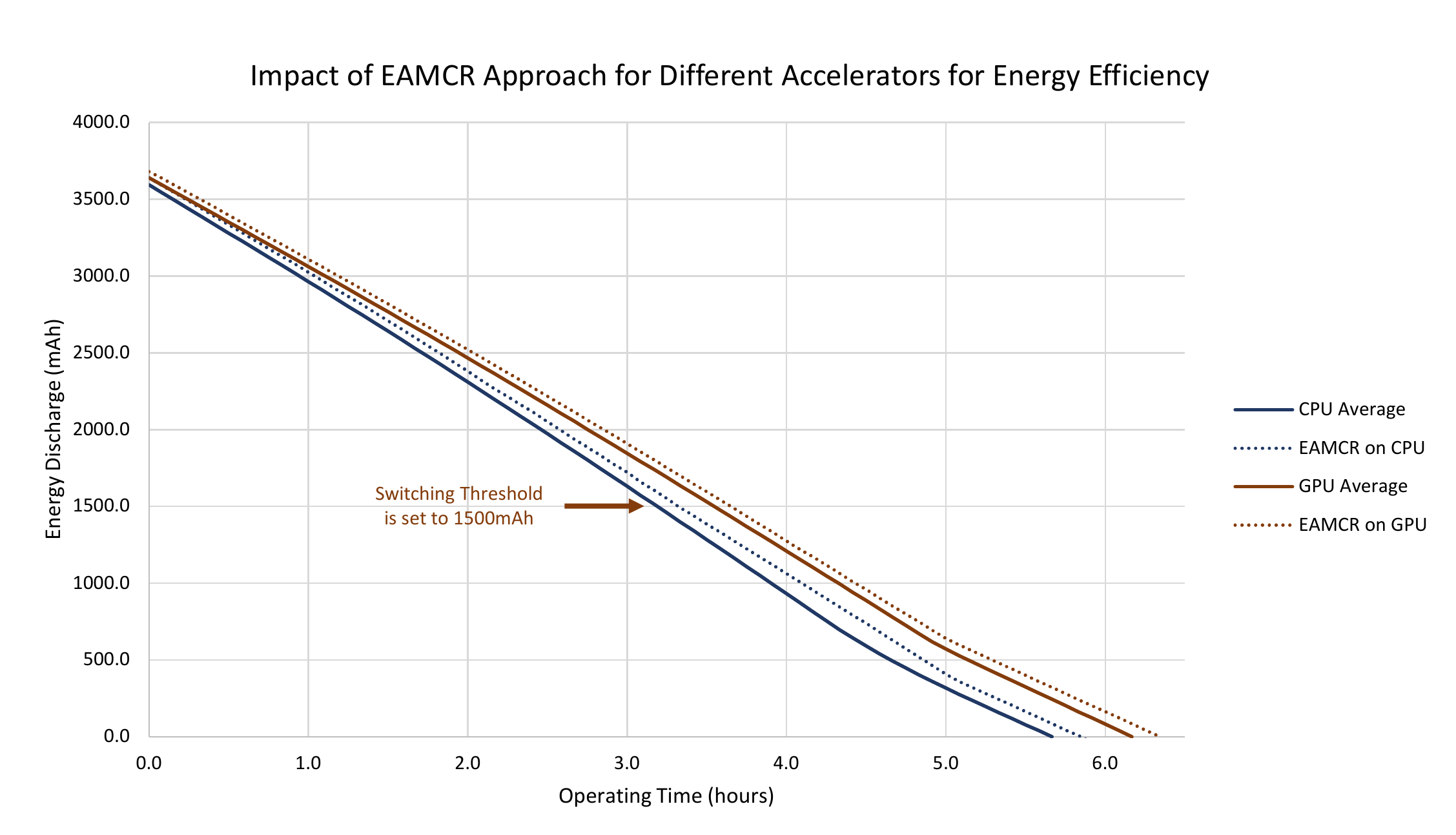}
	\caption{Impact of EAMCR Approach on Different Accelerators.}
	\label{fig:eamcr approach}
\end{figure}

Next, we investigated the behavior of on-device EAMCR approach devised on the same evaluation framework. For this evaluation also, a similar experimental approach is utilized with a similar set of device settings as stated earlier. In addition, a switching threshold of 1500mAh is used and at the startup, the experiments were launched with the highest accurate model for skin lesion diagnosis, which is Inception. The summary of results is shown in the Fig. \ref{fig:eamcr approach} in contrast to the average results of the different accelerators. 

As the results designate, the decision engine starts deep inference with the highest accurate model (in this case Inception) and whenever the switching threshold is reached (in this case 1500mAh), the next best energy efficient model is automatically chosen by the decision engine based on the deployed accelerator. Compared to the average results of all other models, we observed that EAMCR approach always performs better providing higher operating time, regardless of the deployed accelerator.

	\vspace*{-3mm}
\section{Conclusion}
Over the past, on-device deep learning has shown a great potential for enabling mobile-based data-driven applications. However, efficiently deploying deep learning on mobile devices is a kind of balancing act between mobile resource utilization and managing the utility of the intended inference task. In this paper, we investigated the capabilities of deploying current state-of-the art deep learning models and conventional machine learning approaches in different smart health applications through a timely review of recent advancements in on-device deep learning. Moreover, we explored the possibility of adopting crucial model parameters and hardware/software accelerators for deploying energy-aware deep learning framework toward the practical end. Finally, we identified an initial framework (EAMCR) which can be adopted into on-device deep learning in resource contested environments through an energy-aware, adaptive model comprehension and realization approach that utilizes energy characteristics of the underlying DNN.

	\section*{Acknowledgment}
	Effort sponsored in whole or in part by United States Special Operations Command (USSOCOM), under Partnership Intermediary Agreement No. H92222-15-3-0001-01. The U.S. Government is authorized to reproduce and distribute reprints for Government purposes notwithstanding any copyright notation thereon\footnote{The views and conclusions contained herein are those of the authors and should not be interpreted as necessarily representing the official policies or endorsements, either expressed or implied, of the United States Special Operations Command.}.
	
	\bibliographystyle{ACM-Reference-Format}
	\bibliography{articlebibfile}
	
\end{document}